\documentclass[runningheads]{llncs}
\usepackage[T1]{fontenc}

\let\llncsvec\vec
\usepackage{amsmath, amssymb}
\let\vec\llncsvec

\usepackage{comment}
\usepackage{subcaption}
\usepackage{graphicx}
\usepackage{tikz}
\usepackage{xurl}
\usepackage{hyperref}

\usepackage{microtype}
\usepackage{placeins}

\begin{document}

\title{A Causal Argumentation Method for Explainability of Machine Learning Models}
\titlerunning{A Causal-Argumentation Method}

\author{Henry Salgado\inst{1}\orcidID{0009-0001-6506-5484} \and
Meagan R. Kendall\inst{2}\orcidID{0000-0002-9940-4405} \and
Martine Ceberio\inst{1}\orcidID{0000-0001-5680-1155}}

\authorrunning{H. Salgado et al.}

\institute{Department of Computer Science, The University of Texas at El Paso, El Paso, TX, USA \and
Department of Engineering Education and Leadership, The University of Texas at El Paso, El Paso, TX, USA}

\maketitle

\begin{abstract}
Explainable AI (XAI) methods identify which features are relevant to a model's predictions but often fail to clarify \emph{why} certain decisions are made. In this work, we present a novel method that integrates causality with argument-based reasoning to explain why models may be making predictions. Our approach first identifies causal relationships among variables using causal discovery methods and then translates these into a Bipolar Argumentation Framework (BAF) to represent supportive and opposing interactions among features. By using semi-stable semantics, we find extensions of features that explain why certain outcomes may have been chosen. We demonstrate our method on two benchmark datasets and compare its results against standard post-hoc explainability approaches.
\end{abstract}

\section{Introduction}

Machine learning models have become embedded in multiple aspects of societal life. For example, they power the recommendation systems that people interact with daily, the self-driving taxis emerging in cities worldwide, and the productivity tools that shape modern work practices. Although these technologies bring efficiency and innovation, they also raise critical concerns regarding their broader societal implications \cite{christian_alignment_2020}. Indeed, not all applications carry equivalent consequences, and all can have a dramatic negative impact on people through the decisions these models generate. Consider, for example, a model used to determine skin cancer \cite{vedaldi_determining_2020}. How can one ensure that the model's decisions rely on relevant features (i.e., those known by dermatologists), rather than on spurious correlations, such as text on medical images? 

Addressing such questions requires understanding not only \emph{what} a model predicts, but \emph{why} it reaches its conclusions and \emph{how} features causally influence outcomes \cite{pearl_book_2018}. This need for explanation reflects a fundamental human tendency to seek causal understanding of phenomena \cite{miller_explanation_2019}. Humans naturally ask "why" questions: Why did this patient receive this diagnosis? Why was this loan denied? As such, satisfactory answers must go beyond associations to identify causal relationships, that is - the features that, if changed, would alter the outcome. 

Moreover, human reasoning and decision-making are inherently argumentative in nature \cite{beckers_causal_2022}. In courtrooms, lawyers construct arguments from evidence to support or oppose claims. In medicine, physicians weigh competing diagnoses by considering symptoms that support or contradict each hypothesis \cite{josephson_abductive_1996}. In everyday life, humans make decisions by mentally debating the pros and cons of different options. Such an argumentative structure, where reasons support or attack conclusions, provides a natural framework for making reasoning transparent and justifiable and one that can be leveraged to understand machine learning models.

In this work, we propose a \textbf{causal-argumentation method} that connects these perspectives. We take inspiration from the work of Bistarelli et al. \cite{bistarelli_arg-xai_2022}; however, our method differs by integrating causal discovery algorithms with argumentation to produce structured, causally-informed explanations. Specifically, we:
                 
  \begin{enumerate}                                                                                                                                                                             
      \item Identify causal relationships among features using a dual-run
            constraint-based discovery strategy;                       
      \item Translate the resulting causal graph into a Bipolar Argumentation Framework (BAF), where positive and negative correlations define support and attack relations, and convert it to a classical argumentation framework (AF) via               composition rules; and                                          
      \item Use semi-stable semantics to find extensions of features that                                                                                                                       
            support or oppose a given outcome.                                                                                                                                                  
  \end{enumerate} 


We demonstrate our method using two benchmark datasets, the Titanic and Pima Diabetes, and compare its explanations with those generated by standard explanation techniques. Because our method operates on purely observational data, the discovered graphs represent partial equivalence classes of causal structures rather than definitive causal graphs. Nevertheless, our results show that the causal-argumentation approach yields interpretable, structurally grounded explanations that highlight feature interactions while maintaining logical coherence. We also note that our method yields both global (via causal discovery graphs) and local explanations (via argumentation and extensions). 

\section{Background and Related Work}

\subsection{Explainable and Interpretable Machine Learning}

Explainable Artificial Intelligence (XAI), also known as Interpretable Machine Learning (IML), has gained significant momentum over the past decade \cite{molnar_interpretable_2019}. Broadly, explainability approaches can be divided into two categories: post-hoc methods and intrinsically interpretable models (for example decision trees and linear regression). Among these, post-hoc techniques dominate current practice due to their model-agnostic nature, allowing explanations to be applied across diverse architectures without modifying model internals \cite{molnar_interpretable_2019}.

\textbf{Scope of Explanation.}
Explainability methods differ in their level of analysis. Local methods, such as Diverse Counterfactual Explanations (DiCE)\cite{mothilal_explaining_2020} or Local Interpretable Model-agnostic Explanations (LIME) \cite{ribeiro_why_2016}, aim to clarify why a specific prediction was made for an individual instance. On the other hand, global methods, such as Partial Dependence Plots (PDP) \cite{friedman_greedy_2001} or Leave-One-Feature-Out (LOFO), describe the overall decision-making behavior of a model by aggregating feature contributions across instances. Many of these methods rely on testing feature sensitivity or computing feature attribution scores with respect to model outputs.

\textbf{Model Dependency.}
Another key distinction concerns model dependency. Model-agnostic methods (for example SHapley Additive exPlanations (SHAP) and LIME) can be applied to any predictive model, whereas model-specific methods (for example saliency maps for neural networks) exploit the internal structure of the model to derive explanations. In the latter case, explanations are often obtained by inspecting components such as gradients or activations within the kernels of convolutional neural networks. While these approaches enhance transparency, they primarily capture correlations rather than causal or dialectical relationships among features \cite{molnar_general_2022}. 

\subsection{Causal Discovery Methods}

Causality concerns understanding how variables relate, mediate, and interact to explain \emph{why} outcomes occur \cite{molak_causal_2023}. Practitioners often encode causal hypotheses using graphical models that specify directional relationships, for example, \( X \rightarrow Y \) denotes that \( X \) causes \( Y \). Such graphs not only indicate directionality but also determine which causal effects can be reliably estimated \cite{spirtes_causation_2001}. While some graphs are crafted using expert knowledge, algorithmic approaches known as \emph{causal discovery methods} can infer causal structure directly from data under certain assumptions \cite{glymour_review_2019}.

Two major families of causal discovery methods exist: (i) \emph{constraint-based methods} \cite{spirtes_causation_2001} and (ii) \emph{noise-based} (or functional) methods \cite{bystrova_causal_2024}. Noise-based approaches assume specific functional relationships with independent noise terms. For instance, LiNGAM models each variable as a linear combination of its causes plus independent noise. Under this assumption, the statistical properties of \( X \rightarrow Y \) differ from those of \( Y \rightarrow X \), making causal direction identifiable: reversing the direction typically violates the independence between the function and its noise term. While effective for continuous data, these methods rely on strong distributional assumptions such as non-Gaussianity or nonlinearity.

Constraint-based methods, in contrast, infer causal structure from statistical dependence relations among variables. In this paper, we leverage this family of methods because they support mixed data types and impose minimal assumptions about functional form~\cite{bystrova_causal_2024}. The key intuition behind these methods is interventional: if \( X \) causes \( Y \), then there exists at least one value \( x \) such that intervening on \( X \) changes the distribution of \( Y \),
\begin{equation}
\exists x \quad \text{s.t.} \quad P\!\left(Y \mid \mathrm{do}(X = x)\right) \neq P(Y).
\end{equation}
Conversely, if
\begin{equation}
\forall x, \quad P\!\left(Y \mid \mathrm{do}(X = x)\right) = P(Y),
\end{equation}
then intervening on \( X \) has no effect on \( Y \), and thus \( X \) does not cause \( Y \).

While we cannot perform interventions in purely observational settings, constraint-based methods exploit a fundamental insight: 
conditional independence patterns in observational data reflect the underlying causal structure. By systematically testing which variables are (conditionally) independent, these methods reconstruct the graph that would generate those independence patterns. When multiple variables are considered, causal graphs encode \emph{conditional independence} relationships through a concept called \emph{d-separation} (directional separation). Intuitively, if all paths between \( X \) and \( Y \) are ``blocked'' by conditioning on a set \( \mathbf{Z} \), then \( X \) and \( Y \) are d-separated by \( \mathbf{Z} \) and thus conditionally independent. For a set of variables \( \mathcal{V} = \{X_1,\ldots,X_n\} \) and a directed acyclic graph \( \mathcal{G} \), if a set \( \mathbf{Z} \subseteq \mathcal{V} \) d-separates \( X \) and \( Y \) in \( \mathcal{G} \), then
\[
X \perp\!\!\!\perp Y \mid \mathbf{Z}.
\]
For example, consider \textit{education}, \textit{income}, and \textit{health outcomes}. If income d-separates education from health outcomes (meaning education affects health primarily through income), then education and health outcomes are conditionally independent given income: \( \text{education} \perp\!\!\!\perp \text{health} \mid \text{income} \). This corresponds to the causal chain \( \text{education} \rightarrow \text{income} \rightarrow \text{health} \).

Constraint-based causal discovery relies on three key assumptions:
\begin{itemize}
    \item \textbf{Causal Markov Condition:} each variable is independent of its non-descendants given its parents.
    \item \textbf{Faithfulness:} observed independencies in the data correspond exactly to d-separations in the true graph (no independence arises by coincidental parameter cancellation).
    \item \textbf{Causal Sufficiency:} all common causes of observed variables are included in the analysis (no hidden confounders).
\end{itemize}

These assumptions are standard in the causal discovery literature but deserve critical examination, particularly in applied settings. Faithfulness can be violated when parameters in the data-generating process cancel out, producing conditional independencies that do not reflect d-separation in the true graph; while such cancellations are considered measure-zero events in theory~\cite{spirtes_causation_2001}, they may arise approximately in finite samples, especially with modest dataset sizes. Causal sufficiency is difficult to guarantee in practice: unobserved confounders may induce spurious edges or prevent correct orientation. For these reasons, constraint-based methods do not recover a unique causal graph but rather a \emph{Markov equivalence class}, a set of graphs that encode the same conditional independence relations. The output is a partial equivalence class represented as a partially directed graph, where directed edges indicate orientations shared by all members of the class and undirected edges indicate ambiguity that cannot be resolved from observational data alone. We adopt the FCI algorithm precisely because it relaxes the causal sufficiency assumption by allowing for latent confounders, producing a Partial Ancestral Graph (PAG) that represents a more conservative equivalence class. Nevertheless, the resulting structures should be interpreted as \emph{candidate} causal hypotheses consistent with the observed data rather than as confirmed causal relationships. We revisit these limitations in the context of our experimental datasets in Section~4.

Under these assumptions, algorithms such as Peter-Clark (PC) and Fast Causal Inference (FCI) \cite{spirtes_causation_2001} systematically infer causal structure from observational data. Both algorithms proceed in two phases: first identifying adjacencies via conditional independence tests, then orienting edges using structural rules.

The skeleton-building phase starts with a fully connected graph and progressively removes edges. For each pair of variables, the algorithm tests whether they are conditionally independent given various subsets of other variables. If \( X \perp\!\!\!\perp Y \mid \mathbf{Z} \) is found for some conditioning set \( \mathbf{Z} \), the edge between \( X \) and \( Y \) is removed. For instance, suppose we observe that \( A \) and \( C \) are marginally dependent but become independent when conditioning on \( B \):(\( A \perp\!\!\!\perp C \mid B \)). The algorithm would remove the direct edge between \( A \) and \( C \), indicating that their association is entirely mediated through \( B \).

\begin{figure}[htbp]
\centering
\begin{tikzpicture}[node distance=2cm]
    \node (A) {$A$};
    \node (B) [right of=A] {$B$};
    \node (C) [right of=B] {$C$};
    
    \draw[-, thick] (A) -- (B);
    \draw[-, thick] (B) -- (C);
\end{tikzpicture}
\caption{Skeleton after removing the $A$--$C$ edge based on $A \perp\!\!\!\perp C \mid B$.}
\label{fig:skeleton}
\end{figure}
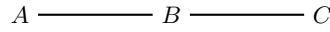

The edge-orientation phase then uses distinctive statistical patterns to determine causal direction where possible. A critical orientation pattern is the \emph{unshielded collider} (or v-structure): \( X \rightarrow Y \leftarrow Z \), where two non-adjacent variables \( X \) and \( Z \) both influence a common effect \( Y \). Colliders are uniquely identifiable because they create a distinctive statistical signature: the causes are independent until we condition on their common effect, at which point they become dependent through a phenomenon called ``explaining away.'' This pattern cannot arise from other causal structures. For example, if \( A \) and \( C \) are marginally independent but both connect to \( B \) (with no \( A \)--\( C \) edge remaining from the skeleton phase), the structure must be \( A \rightarrow B \leftarrow C \). Once colliders are oriented, propagation rules orient remaining edges while avoiding the introduction of new colliders or cycles.

\begin{figure}[htbp]
\centering
\begin{tikzpicture}[node distance=2cm]
    \node (A) {$A$};
    \node (B) [right of=A] {$B$};
    \node (C) [right of=B] {$C$};
    
    \draw[->, thick] (A) -- (B);
    \draw[->, thick] (C) -- (B);
\end{tikzpicture}
\caption{Oriented collider structure $A \rightarrow B \leftarrow C$ identified from the pattern where $A$ and $C$ are marginally independent but both adjacent to $B$.}
\label{fig:collider}
\end{figure}
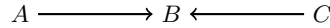

It is important to note that these methods often cannot orient all edges from observational data alone, resulting in partially oriented graphs where some edges remain undirected when orientation cannot be determined. PC assumes causal sufficiency and outputs a Partially Directed Acyclic Graph (PDAG), whereas FCI relaxes this assumption by introducing bidirected edges (\( X \leftrightarrow Y \)) to represent possible latent confounders, outputting a PAG that provides a more conservative representation when hidden variables may be present.

\subsection{Argumentation Frameworks}

In life, we often need to reason about sets of features that collectively support or contradict a decision or prediction. For instance, when explaining a medical diagnosis, some patient characteristics may reinforce the prediction while others conflict with it. Abstract Argumentation Frameworks (AFs) \cite{dung_acceptability_1995} provide a formal model for reasoning about such conflicting and supporting information through the interaction of arguments.

In an AF, the acceptability of an argument depends not on its intrinsic strength but on how it relates to others, specifically, on the patterns of \emph{attack} and \emph{defense} among arguments. Formally, an AF is defined as a pair \(\mathcal{F} = (\mathcal{A}, \mathcal{R})\), where \(\mathcal{A}\) is a set of arguments and \(\mathcal{R} \subseteq \mathcal{A} \times \mathcal{A}\) is a binary relation representing attacks between arguments. If \((a,b) \in \mathcal{R}\), we say that argument \(a\) attacks \(b\).

For example, consider three arguments about a patient: \(a_1 =\) ``high blood pressure'', \(a_2 =\) ``young age'', and \(a_3 =\) ``sedentary lifestyle''. We might have \(a_2\) attacks \(a_1\) (young age argues against high blood pressure being a primary concern), while \(a_3\) reinforces \(a_1\) (sedentary lifestyle supports the concern about high blood pressure). This intuition motivates the need for frameworks that can represent both opposition and support.

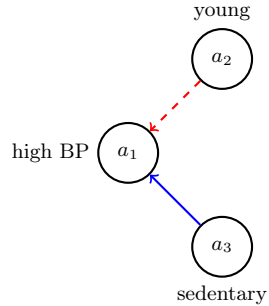
\begin{figure}[htbp]
\centering
\begin{tikzpicture}[
    scale=0.88,
    transform shape,
    node distance=2.0cm,
    argument/.style={circle, draw, minimum size=0.9cm, thick}
]
    \node[argument, label={[font=\small]left:{high BP}}] (a1) {$a_1$};
    \node[argument, label={[font=\small]above:{young}}] (a2) [above right of=a1] {$a_2$};
    \node[argument, label={[font=\small]below:{sedentary}}] (a3) [below right of=a1] {$a_3$};

    \draw[->, thick, dashed, red] (a2) -- (a1);

    \draw[->, thick, blue] (a3) -- (a1);
\end{tikzpicture}
\caption{Example bipolar argumentation framework showing attack (dashed red) and support (solid blue) relations.}
\label{fig:baf_example}
\end{figure}

Bipolar Argumentation Frameworks (BAFs) \cite{cayrol_acceptability_2005} extend the standard AF model by explicitly introducing both \emph{support} and \emph{attack} relations among arguments. In a BAF, the structure is defined as \(\mathcal{F} = (\mathcal{A}, \mathcal{R}^+, \mathcal{R}^-)\), where \(\mathcal{R}^+ \subseteq \mathcal{A} \times \mathcal{A}\) denotes supportive relations and \(\mathcal{R}^- \subseteq \mathcal{A} \times \mathcal{A}\) denotes attacks. For example, if \(x, y \in \mathcal{A}\) and \((x, y) \in \mathcal{R}^+\), then \(x\) supports \(y\); conversely, \((x, y) \in \mathcal{R}^-\) indicates that \(x\) attacks \(y\). This dual representation allows argumentation systems to capture both reinforcing and opposing relationships, providing a richer model of reasoning than standard AFs.

Furthermore, given an argumentation framework, we can determine which sets of arguments are \emph{acceptable} under a chosen semantics. Different semantics provide different criteria for what constitutes a coherent set of arguments. A set \(S \subseteq \mathcal{A}\) is said to be \emph{conflict-free} if no two arguments in \(S\) attack each other, i.e., there are no \(a, b \in S\) such that \((a,b) \in \mathcal{R}^-\). A set \(S\) is \emph{admissible} if it is conflict-free and defends all of its members—that is, for every argument \(c \notin S\) that attacks some \(a \in S\), there exists another argument \(b \in S\) that attacks \(c\). Intuitively, an admissible set is internally consistent and can counter all external attacks.

Among the various semantics proposed for AFs, \emph{semi-stable semantics} \cite{caminada_semi-stable_2006} strike a balance between consistency and expressiveness. To understand semi-stable semantics, we first define the \emph{range} of a set \(S\), which includes both the arguments in \(S\) and all arguments attacked by \(S\):
\[
S^+ = S \cup \{a \in \mathcal{A} \mid \exists b \in S : (b,a) \in \mathcal{R}^-\}.
\]
Intuitively, \(S^+\) represents the ``sphere of influence'' of \(S\), the arguments that are either accepted or explicitly rejected by \(S\). A semi-stable extension is then an admissible set that maximizes this range. Unlike stable semantics, which require every argument outside \(S\) to be attacked (and may not exist for all frameworks), semi-stable extensions are guaranteed to exist and capture the broadest set of arguments that can be coherently decided.

We select semi-stable semantics for three reasons that are particularly relevant to our explainability setting. First, the argumentation frameworks constructed by our method are always finite, since they are derived from a finite number of discretized feature instances in the dataset. For finite AFs, semi-stable extensions are guaranteed to exist~\cite{caminada_semi-stable_2006}, whereas stable extensions may fail to exist for certain attack structures. Second, semi-stable semantics maximize the range $S^+$, which means they take a position, either accepting or explicitly rejecting, on as many feature arguments as possible. In an explainability context, this is desirable because it produces explanations that address the broadest set of features rather than leaving many features undecided. Third, compared to preferred semantics (which are also guaranteed to exist for finite AFs), semi-stable extensions tend to yield more decisive results: preferred extensions may leave large portions of the framework unresolved, which would produce incomplete explanations. In summary, semi-stable semantics occupy a useful middle ground between stable semantics (maximally decisive but not guaranteed to exist) and preferred semantics (always exist but potentially leave too many arguments undecided), making them well-suited for generating comprehensive feature-based explanations.

\section{Methods}

\subsection{Overview}

We now describe our proposed \emph{causal-argumentation method}, which connects causal discovery with argumentative reasoning. The method proceeds through four main stages: (1) data preparation, (2) causal discovery, (3) unified feature encoding, and (4) construction of a BAF with probabilistic reasoning under semi-stable semantics. The complete implementation is available at \url{https://github.com/eigenhenry/causal_argumentation/tree/main}.

\subsection{Data Preparation and Feature Encoding}

To capture more nuanced patterns among features, we apply transformations prior to causal discovery. Rather than only testing whether a continuous variable such as \textit{Age} is relevant, we seek to identify specific value ranges that may meaningfully influence the outcome (for example, \textit{Age} < 5 representing young passengers). Similarly, for categorical variables such as \textit{Sex}, we consider distinct instances (for example, \textit{Sex} = male, \textit{Sex} = female) to preserve interpretive clarity in subsequent argumentative reasoning.

Numerical features are discretized using \emph{supervised entropy-based binning}, partitioning values to maximize information gain with respect to the outcome. The number of bins can be adjusted depending on domain knowledge about the data. For example, we might discretize \textit{Age} into three bins representing child, adult, and elderly, or use finer-grained intervals when the data suggests more specific age-related patterns. Depending on the feature context, categorical variables are either one-hot encoded or binarized to indicate the presence or absence of an attribute. For example, variables such as \textit{Sex} are best represented through one-hot encoding to retain category identity, whereas count-based variables such as \textit{SibSp} (number of siblings or spouses aboard) can be binarized (for example, at least one versus none) or discretized using entropy-based thresholds when more nuanced distinctions are needed.

\subsection{Dual-Run Causal Discovery}
We follow the methodology detailed in our previous work \cite{salgado_causal_2026}, which addresses the challenge of singular or ill-conditioned covariance matrices that arise when categorical variables are fully one-hot encoded in causal discovery. The approach implements a \emph{dual-run strategy} using complementary encoding schemes: \textit{drop-first} and \textit{drop-last}. Each encoding removes a different dummy variable from each categorical group, ensuring the covariance matrix remains invertible for conditional independence tests.

A key component of our approach is enforcing the outcome variable as a sink node, meaning all directed paths must terminate at the target (e.g., \textit{Survived} in the Titanic dataset). This constraint encodes the domain knowledge that outcome variables do not causally influence predictor values in observational settings.

We apply the FCI algorithm independently to each encoding variant using the \texttt{causal-learn} library \cite{zheng_causal-learn_2023} with significance level \(\alpha=0.01\) and Fisher's \(z\)-test. The FCI algorithm handles potential unmeasured confounding by leaving certain edges unoriented when causal direction cannot be reliably determined from the observed data. Each run produces a PAG representing the discovered causal structure.

\subsection{Unified Encoded Graph Construction}
Following the dual-run approach, we merge the two encoding-specific PAGs using majority voting. An edge is retained only when one orientation strictly dominates: \(A \rightarrow B\) is included in the merged graph if and only if it appears in strictly more runs than \(B \rightarrow A\). When both runs produce conflicting orientations for the same variable pair, the edge is excluded from the merged graph entirely. 

The workflow produces graphs at two levels:
\begin{itemize}
    \item \textbf{Discovery graphs:} encoding-specific PAGs from drop-first and drop-last runs
    \item \textbf{Unified graph:} merged representation with correlation-weighted edges
\end{itemize}

\subsection{Bipolar Argumentation Framework (BAF)}

From the unified graph, we construct a BAF \(B = (A, R^+, R^-)\), where arguments \(A\) represent feature instances, and edges define support (\(R^+\)) or attack (\(R^-\)) relations based on Pearson correlation coefficients. Additionally, we introduce a negative correlation (\(-1\)) between feature instances derived from the same original variable (for example, \textit{Age[0,5)} versus \textit{Age[5,inf)}). This constraint ensures that complementary intervals or categories are treated as mutually exclusive within the BAF, preventing logically incompatible feature instances from being simultaneously included in the same argumentative extension. The outcome variable is split into two exclusive arguments (for example \texttt{survived\_0} and \texttt{survived\_1}) to capture alternative hypotheses.

We note that the resulting argumentation framework is always finite, since the set of arguments is bounded by the number of discretized feature instances and outcome categories derived from a finite dataset. This finiteness guarantees the existence of semi-stable extensions~\cite{caminada_semi-stable_2006}.

\subsection{Conversion to Classical AF and Probabilistic Reasoning}

The BAF is converted to a classical AF \cite{dung_acceptability_1995} by applying                                                                                                             
  three composition rules. First, \textit{support closure}: if \(A\) supports \(B\)                                                                                                             
  and \(B\) supports \(C\), a new support \(A \rightarrow C\) is derived with weight                                                                                                            
  \(w_{AB} \cdot w_{BC}\), applied iteratively until fixpoint. Second,                                                                                                                          
  \textit{support-mediated attack}: if \(A\) supports \(B\) and \(B\) attacks \(C\),
  then \(A\) is derived to attack \(C\) with weight \(w_{AB} \cdot w_{BC}\). Third,                                                                                                             
  \textit{defender counter-attack}: if \(A\) attacks \(B\) and \(C\) supports \(B\),                                                                                                            
  then \(C\) is derived to attack \(A\) with weight \(w_{CB}\), reflecting that a                                                                                                               
  supporter of \(B\) opposes \(B\)'s attacker. Composed relations with weight below                                                                                                             
  a threshold \(\tau\) are discarded.    

Additionally, weak correlations (\(|r| < 0.05\)) are pruned before transforming the BAF into a classical AF \cite{dung_acceptability_1995}. Attacks are treated as probabilistic events with inclusion likelihood proportional to correlation magnitude. Using a constellation approach, we generate the top \(k=50\) possible worlds and compute semi-stable extensions.

\FloatBarrier
\section{Evaluation and Discussion}


We evaluate our approach on two benchmark binary classification datasets, comparing interpretability and alignment with feature importance scores produced by SHAP, LIME, and pruned decision trees. The results demonstrate that the proposed causal-argumentation framework preserves the directional structure discovered from observational data while providing human-readable explanations of feature interactions. As noted in Section~2.2, the discovered graphs represent partial equivalence classes rather than definitive causal structures, so the explanations produced by our method should be understood as structurally grounded hypotheses consistent with the observed data.

\subsection{Titanic Dataset}

The Titanic dataset contains information about 891 passengers with features including \textbf{Pclass} (passenger class), \textbf{Sex}, \textbf{Age}, \textbf{SibSp} (siblings/spouses aboard), \textbf{Parch} (parents/children aboard), \textbf{Fare}, and \textbf{Embarked} (port), predicting survival (\(y=1\)) or death (\(y=0\)).

We discretized numerical variables using entropy-based binning and one-hot encoded categorical variables. The FCI algorithm was executed twice (\(\alpha=0.01\)) using drop-first and drop-last encoding strategies with Fisher's \(z\)-test for conditional independence (though $G^2$ or kernel-based tests can substitute based on data properties). The graphs were merged through majority voting to form a unified causal structure.

\begin{figure}[htbp]
    \centering

    \begin{subfigure}{0.70\textwidth}
        \centering
        \includegraphics[width=\textwidth]{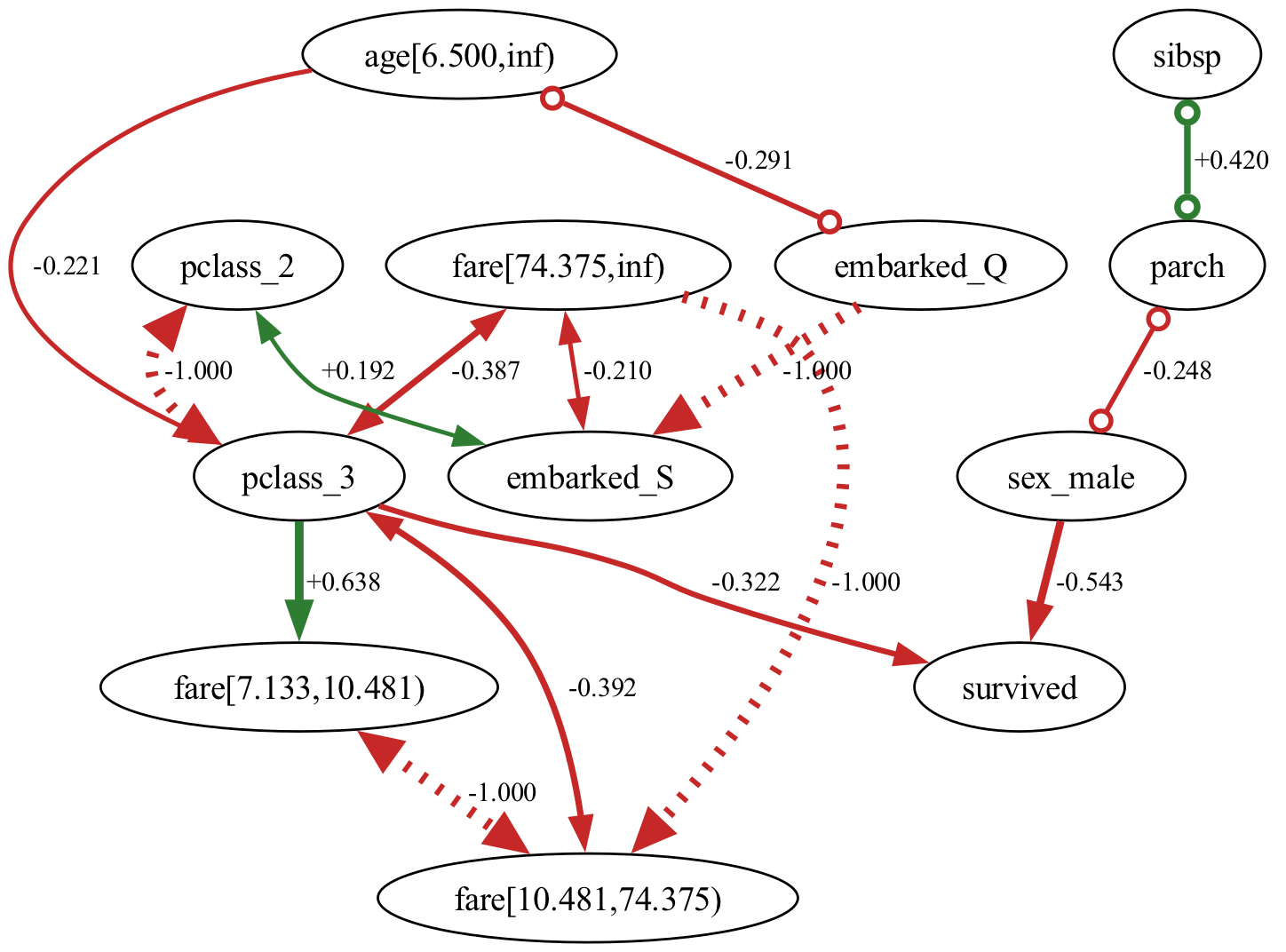}
        \caption{Causal graph using drop-first encoding.}
        \label{fig:dropfirst_graph}
    \end{subfigure}

    \vspace{1em}

    \begin{subfigure}{0.7\textwidth}
        \centering
        \includegraphics[width=\textwidth]{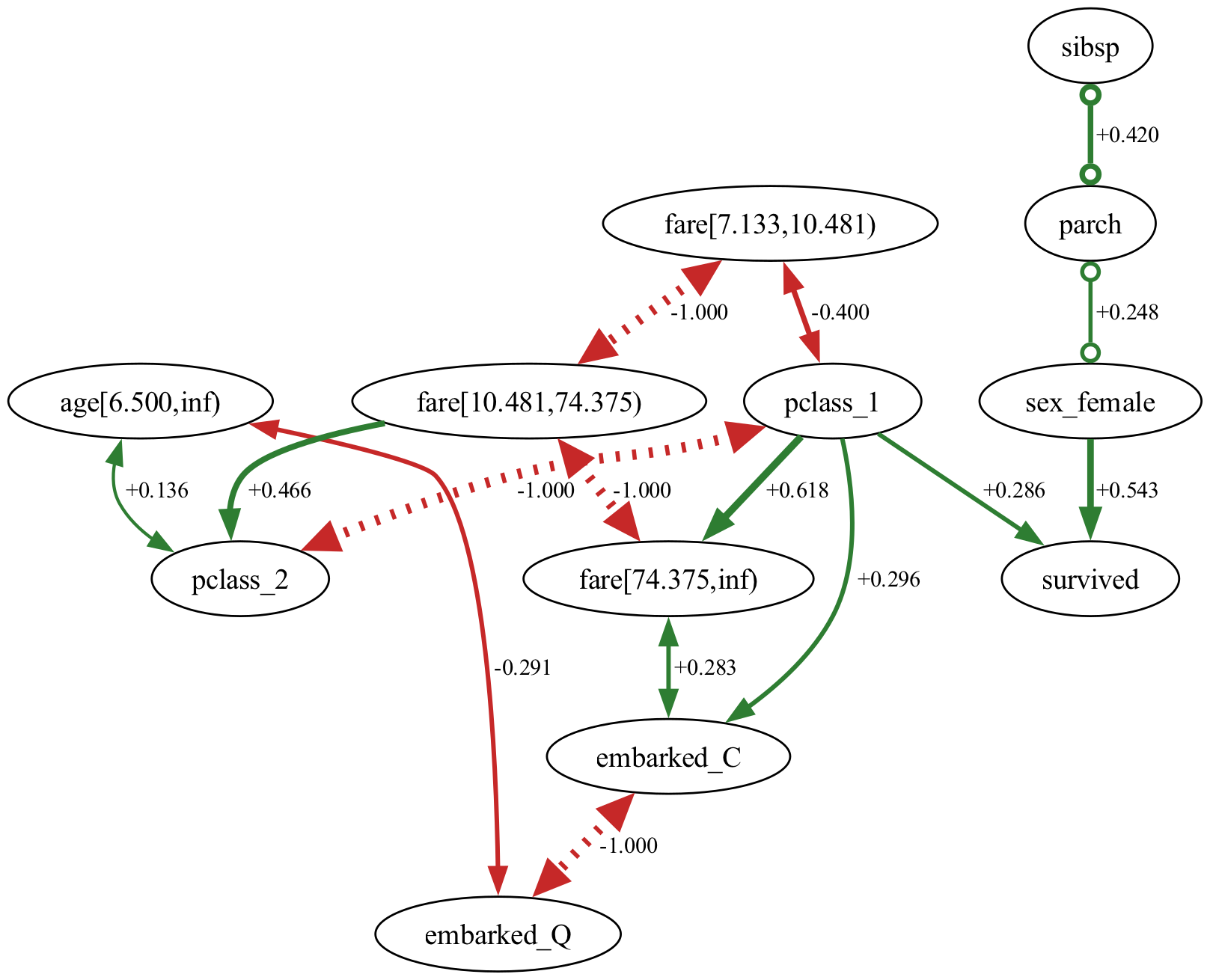}
        \caption{Causal graph using drop-last encoding.}
        \label{fig:droplast_graph}
    \end{subfigure}

    \caption{Comparison of causal graphs generated under two encoding strategies for the Titanic dataset. The structural consistency across both encodings reflects robustness in edge orientation and causal directionality.}
    \label{fig:titanic_encoding_comparison}
\end{figure}

Both encoding strategies produced consistent causal orientations (Fig.~\ref{fig:titanic_encoding_comparison}). \textit{Sex} directly influences \textit{Survived}, with strong negative correlation for males. Higher passenger class (\textit{Pclass\_1}) positively influences survival, while lower class (\textit{Pclass\_3}) shows negative influence. \textit{Fare} emerged as a mediating factor between class and survival, with lower intervals negatively associated and higher intervals positively associated with survival. \textit{Embarked} variables connected indirectly to class and fare, suggesting location-based socioeconomic differences. Sibling and parent variables showed weak, localized effects.

\FloatBarrier

The unified graph translated into a BAF with \(|A|=17\) arguments, \(|R^+|=15\) supportive relations, and \(|R^-|=17\) attacks. After conversion to classical AF, the structure contained 18 arguments and 57 attack relations. Under semi-stable semantics, two representative extensions emerged:
\[
\begin{aligned}
E_1 &= \{\textit{age[-inf,6.500)}, \textit{embarked\_C}, \textit{fare[-inf,7.133)}, \textit{parch},\\
    &\quad \textit{pclass\_3}, \textit{sex\_male}, \textit{sibsp}, \textit{survived\_0}\},\\[4pt]
E_2 &= \{\textit{age[-inf,6.500)}, \textit{embarked\_C}, \textit{fare[10.481,74.375)}, \textit{parch},\\
    &\quad \textit{pclass\_1}, \textit{sex\_female}, \textit{sibsp}, \textit{survived\_1}\}.
\end{aligned}
\]

\subsubsection{Interpreting Extensions as Argumentative Explanations.}

We now trace the argumentative reasoning that justifies why each set of features is coherent and what it explains about the predicted outcome.

\textbf{Extension \(E_1\) (non-survival).} This extension explains non-survival as follows: \emph{the passenger was male and traveled in third class with a low fare. Being male strongly argues against survival, and third-class status reinforces non-survival through the low fare pathway. Being female and being in first class, which would argue for survival, are excluded because they directly conflict with being male and being in third class, respectively.}

\textbf{Extension \(E_2\) (survival).} This extension explains survival as follows: \emph{the passenger was female and traveled in first class with a moderate fare. Being female strongly supports survival, reflecting the ``women and children first'' evacuation policy, and first-class status reinforces this through the higher fare pathway. Being male and being in third class, which would argue against survival, are excluded because they directly conflict with being female and being in first class, respectively.}

\textbf{Dialectical contrast.} The two extensions represent competing cases for the two possible outcomes. Features accepted in one extension (e.g., being male in \(E_1\)) directly conflict with features accepted in the other (e.g., being female in \(E_2\)). This mirrors how a human decision-maker might weigh evidence: the same set of features can tell two coherent but mutually exclusive stories depending on which values are present. The semi-stable semantics ensure that each story is internally consistent, can defend all of its members against opposing features, and takes a position on as many features as possible.

\subsubsection{Validation Against Baseline Methods.}

\begin{figure}[htbp]
    \centering
    \includegraphics[width=0.8\textwidth]{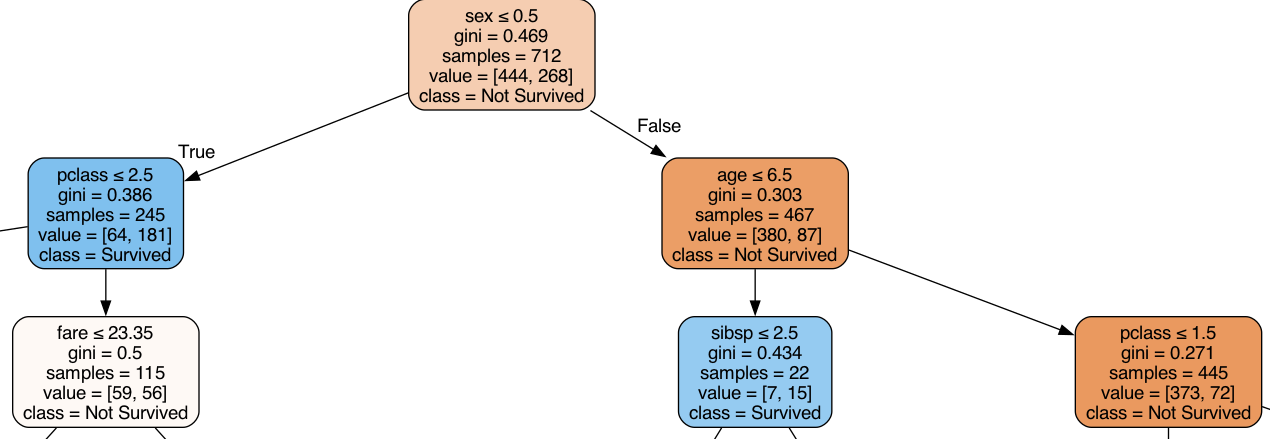}
    \caption{Pruned decision tree trained on the dataset. The root split on \textit{sex}, followed by splits on \textit{pclass} and \textit{age}, mirrors the discriminative structure captured by the semi-stable extensions.}
    \label{fig:decision_tree}
\end{figure}

The pruned decision tree (Fig.~\ref{fig:decision_tree}) exhibits patterns closely paralleling our semi-stable extensions. The root split on \textit{sex} indicates gender as most discriminative, with female passengers showing substantially higher survival frequency, consistent with \textit{sex\_female} prominence in our extensions. Subsequent splits on \textit{pclass} and \textit{age} (approximately 6.5 years) reinforce this alignment.

\FloatBarrier

\begin{figure}[htbp]
    \centering
    \includegraphics[width=0.8\textwidth]{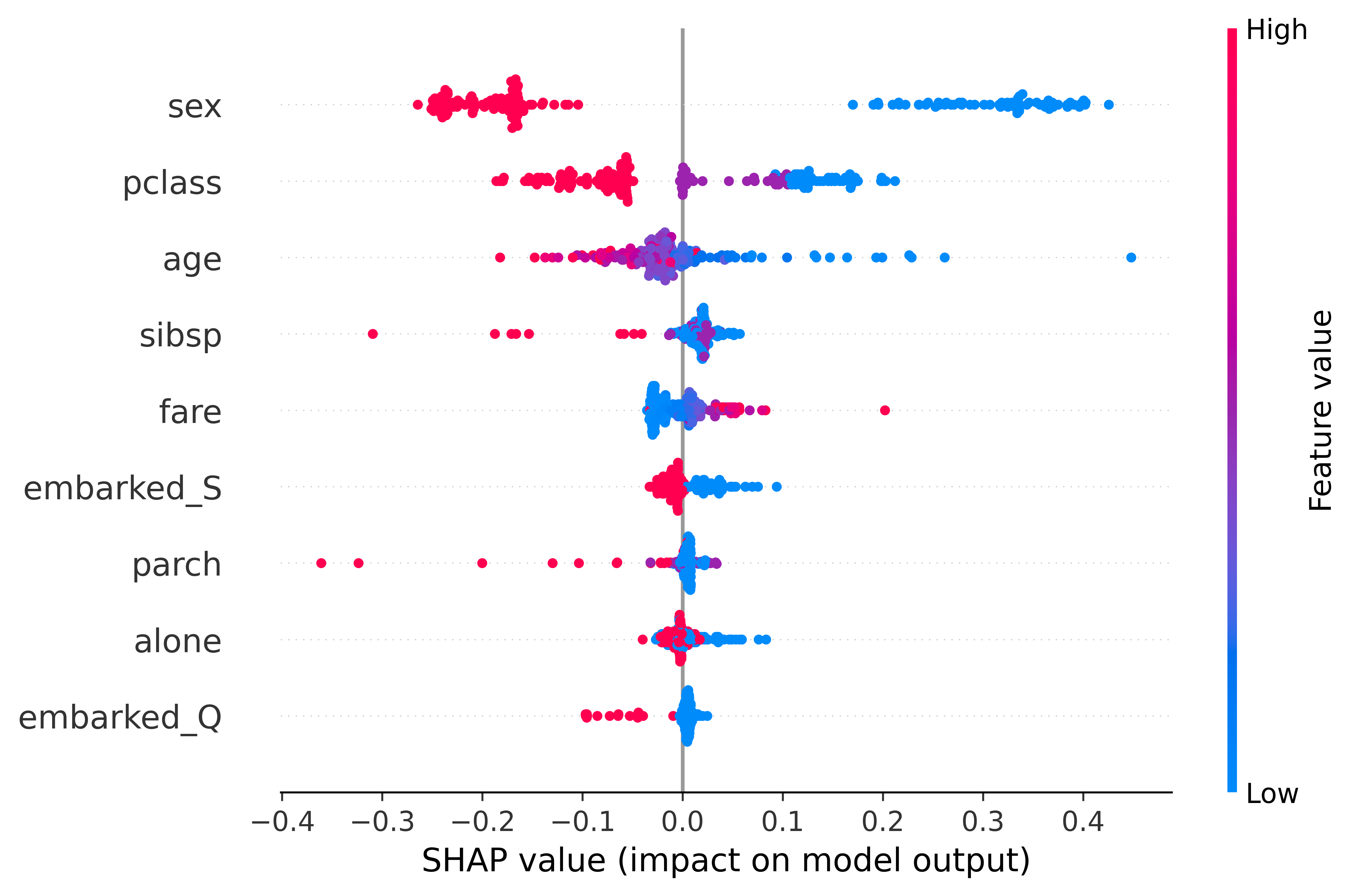}
    \caption{SHAP summary plot illustrating global feature importance. The strongest contributors, \textit{sex}, \textit{pclass}, and \textit{age}, align with the features most frequently retained in the semi-stable extensions.}
    \label{fig:shap_summary}
\end{figure}

SHAP analysis (Fig.~\ref{fig:shap_summary}) corroborates these findings, ranking \textit{sex}, \textit{pclass}, and \textit{age} as strongest contributors. These are the same features most consistently retained across our semi-stable extensions. This convergent validity confirms our framework captures discriminative patterns recognized by established baselines.

\FloatBarrier

\subsection{Pima Indians Diabetes Dataset}

The Pima Indians Diabetes dataset contains medical measurements for 768 female patients with features including \textbf{Pregnancies}, \textbf{Glucose}, \textbf{BloodPressure}, \textbf{SkinThickness}, \textbf{Insulin}, \textbf{BMI}, \textbf{DiabetesPedigreeFunction}, and \textbf{Age}, predicting diabetes diagnosis (\(y=1\)) or no diagnosis (\(y=0\)).

\begin{figure}[htbp]
    \centering
    \includegraphics[width=1\textwidth]{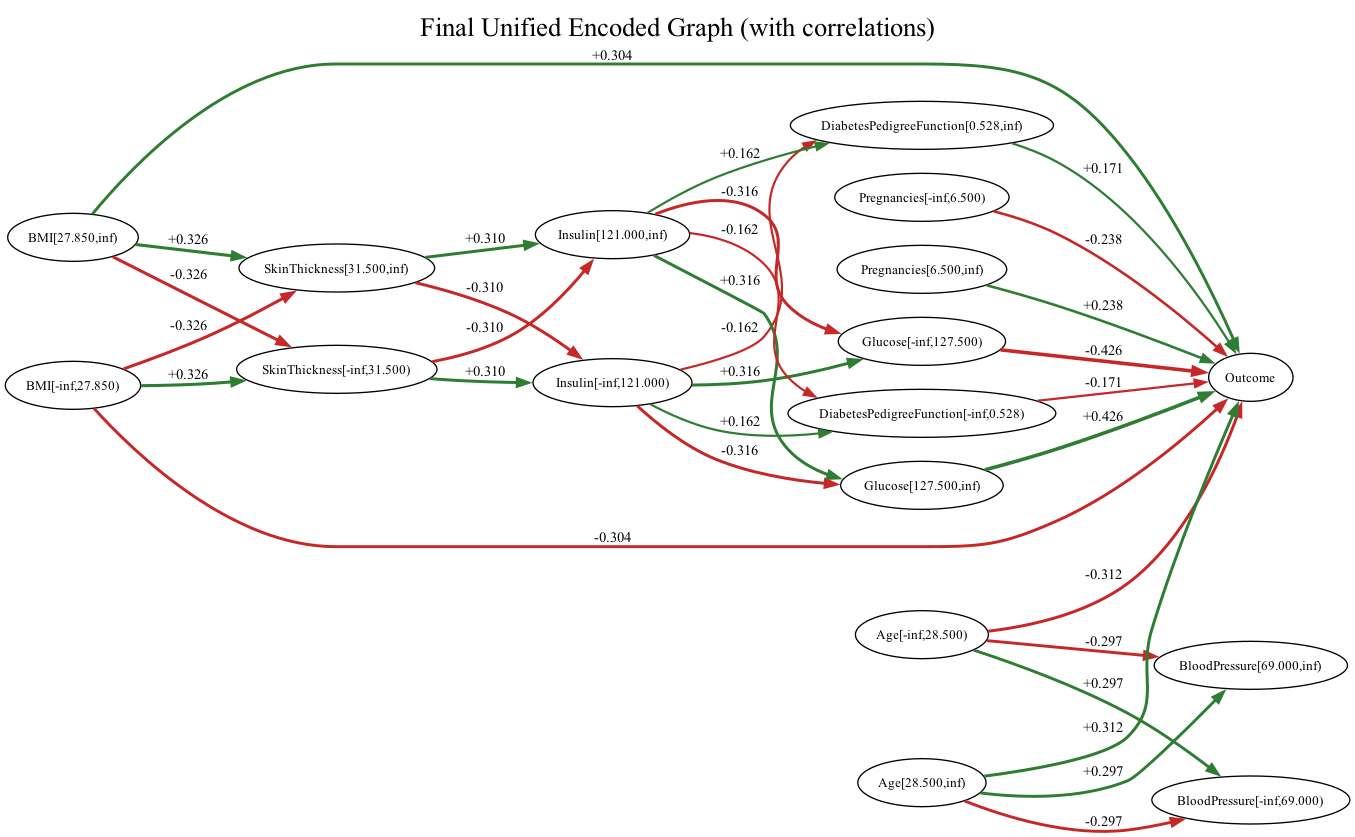}
    \caption{Unified causal graph for the Pima Indians Diabetes dataset, merged from both encoding strategies with correlation weights.}
    \label{fig:diabetes_unified}
\end{figure}

We applied the same pipeline with entropy-based binning and dual-encoding FCI (\(\alpha=0.01\)) using Fisher's \(z\)-test, then merged graphs via majority voting. The unified causal graph (Fig.~\ref{fig:diabetes_unified}) shows consistent orientations across both encoding strategies. \textit{Glucose[127.500,inf)} directly influences diabetes outcome, with support from \textit{Insulin[121.000,inf)} and \textit{BMI[27.850,inf)}. These paths reflect physiological mechanisms: elevated glucose, insulin resistance, and body mass jointly contribute to diabetes likelihood. Additional positive paths from \textit{SkinThickness[31.500,inf)} to \textit{Insulin}, and from \textit{Pregnancies[6.500,inf)} and \textit{Age[28.500,inf)} to \textit{BMI}, reflect demographic and metabolic influences.

\FloatBarrier

The unified graph yielded a BAF with \(|A|=17\) arguments, \(|R^+|=13\) supportive relations, and \(|R^-|=13\) attacks, expanding to 18 arguments and 100 attack relations in classical AF. Two representative semi-stable extensions:
\begin{multline*}
E_1 = \{\textit{Age[28.500,inf)}, \textit{BMI[27.850,inf)}, 
\textit{BloodPressure[-inf,69.000)},\\
\textit{DiabetesPedigreeFunction[0.528,inf)}, 
\textit{Glucose[127.500,inf)}, \textit{Insulin[121.000,inf)},\\
\textit{Outcome\_1}, \textit{Pregnancies[6.500,inf)}, 
\textit{SkinThickness[31.500,inf)}\},\\[4pt]
E_2 = \{\textit{Age[-inf,28.500)}, \textit{BMI[-inf,27.850)}, 
\textit{BloodPressure[-inf,69.000)},\\
\textit{DiabetesPedigreeFunction[-inf,0.528)}, 
\textit{Glucose[-inf,127.500)}, \textit{Insulin[-inf,121.000)},\\
\textit{Outcome\_0}, \textit{Pregnancies[-inf,6.500)}, 
\textit{SkinThickness[-inf,31.500)}\}.
\end{multline*}

\subsubsection{Interpreting Extensions as Argumentative Explanations.}

We again trace the argumentative reasoning that justifies why each set of features is coherent and what it explains about the predicted outcome.

\textbf{Extension \(E_1\) (diabetic profile).} This extension explains a diabetes diagnosis as follows: \emph{the patient has elevated glucose, high insulin, high BMI, older age, multiple pregnancies, greater skin thickness, and a strong family history of diabetes. These features mutually reinforce one another, for example, older age and more pregnancies are associated with higher BMI, and greater skin thickness is associated with higher insulin levels, which in turn is associated with higher glucose. All competing lower-range values (e.g., normal glucose, low BMI) are excluded because they directly conflict with the elevated values in this extension, and no accepted features contradict this conclusion.}

\textbf{Extension \(E_2\) (non-diabetic profile).} This extension tells the opposite story: \emph{the patient has normal glucose, low insulin, healthy BMI, younger age, fewer pregnancies, lower skin thickness, and no strong family history. These features consistently point away from diabetes, and all elevated values from \(E_1\) are excluded because they conflict with the lower-range values accepted here. The result is a coherent profile of a low-risk patient with no features arguing in favor of a diabetes diagnosis.}

\textbf{Dialectical contrast.} As with the Titanic dataset, the two extensions form opposing cases. Each feature accepted in \(E_1\) has a conflicting counterpart accepted in \(E_2\). The relationships identified by the discovery phase (e.g., pregnancies are associated with BMI, which is associated with the outcome) are preserved in both extensions, but the feature values that are present lead to opposite conclusions. This structure makes explicit \emph{why} different feature values lead to different outcomes: the same relationships hold, but the direction of the conclusion reverses depending on where the patient falls within each feature's range.

\subsubsection{Validation Against Baseline Methods.}

\begin{figure}[htbp]
    \centering
    \includegraphics[width=0.9\textwidth]{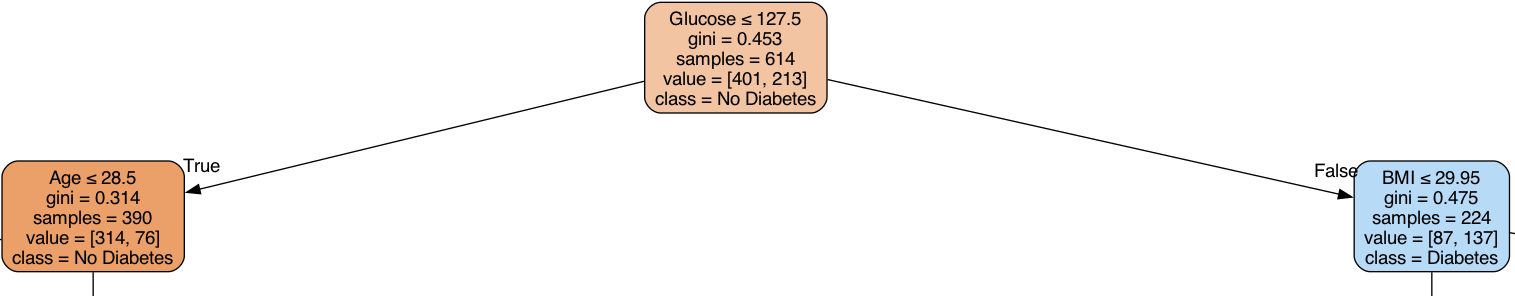}
    \caption{Pruned decision tree trained on the Pima Diabetes dataset. The root split on \textit{Glucose} and subsequent splits on \textit{Age} and \textit{BMI} reflect the discriminative structure captured by the semi-stable extensions.}
    \label{fig:diabetes_tree}
\end{figure}

The decision tree (Fig.~\ref{fig:diabetes_tree}) reveals structures paralleling our extensions. The root split on \textit{Glucose} (\(\approx\)127.5) indicates lower values predict lower diabetes likelihood, consistent with our argumentative findings. Subsequent splits on \textit{Age} (\(\approx\)28.5) and \textit{BMI} (\(\approx\)29.95) mirror the importance assigned to these features in our semi-stable results.

\FloatBarrier

\begin{figure}[htbp]
    \centering
    \includegraphics[width=0.9\textwidth]{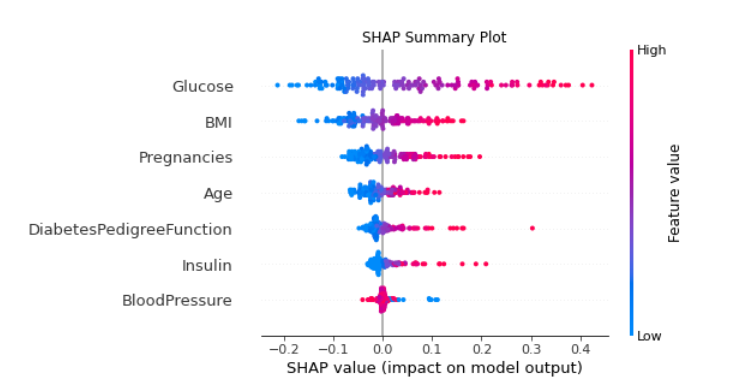}
    \caption{SHAP summary plot for the Pima Diabetes dataset. The strongest contributors, \textit{Glucose}, \textit{BMI}, \textit{Pregnancies}, and \textit{Age}, align with the features emphasized in the semi-stable extensions.}
    \label{fig:diabetes_shap}
\end{figure}

SHAP analysis (Fig.~\ref{fig:diabetes_shap}) identifies \textit{Glucose}, \textit{BMI}, \textit{Pregnancies}, and \textit{Age} as strongest contributors, closely matching arguments most consistently retained in our extensions. This convergence confirms our argumentative approach captures principal discriminative patterns recognized by established baselines.

\FloatBarrier

\section{Limitations and Scope of Causal Claims}

\textbf{Observational setting.} Our analysis is fully observational: no interventional experiments were conducted, and the discovered graphs have not been validated against known ground-truth causal structures. The FCI algorithm outputs a Partial Ancestral Graph (PAG) that represents a Markov equivalence class, meaning multiple distinct causal graphs are consistent with the same set of conditional independence relations in the data. Consequently, the directed edges in our unified graphs should be interpreted as orientations shared across members of this equivalence class, not as confirmed causal directions. Undirected edges in the PAG explicitly mark relationships where causal direction remains ambiguous.

\textbf{Assumption plausibility on the experimental datasets.} The Titanic dataset (891 observations, 7 features) and the Pima Diabetes dataset (768 observations, 8 features) are modest in size, which limits the statistical power of conditional independence tests and increases the risk of both false positives and false negatives in edge detection. Hidden confounding is plausible in both datasets: for example, socioeconomic factors not captured in the Titanic data may jointly influence both passenger class and survival, and lifestyle variables absent from the Pima dataset (such as diet and exercise) may confound relationships among BMI, glucose, and diabetes. While the FCI algorithm is designed to accommodate latent confounders through bidirected edges, it cannot fully recover confounded structures when the confounders are entirely unobserved. Faithfulness violations are also possible in finite samples, where approximate parameter cancellations may produce misleading independence patterns.

\textbf{Scope of our claims.} In light of these considerations, we frame our contribution as follows: the causal discovery step provides a \emph{structurally informed} starting point that goes beyond purely correlational analysis by leveraging conditional independence patterns and directional constraints. The argumentation layer then organizes these discovered relationships into coherent, dialectically structured explanations. The resulting explanations are best understood as plausible hypotheses about why certain predictions arise, grounded in the statistical structure of the data, rather than as confirmed causal accounts. Stronger causal claims would require either interventional validation or the use of datasets with known ground-truth causal graphs, which we identify as an important direction for future work.

\section{Conclusion and Future Work}

In this work, we introduced a novel explainability method that integrates causal discovery with abstract argumentation to generate structured, interpretable insights into machine-learning models. By combining a global causal representation with local argumentation-based reasoning, our approach provides a unified framework capable of capturing both global variable relationships and local, instance-level explanations. Validation on two benchmark datasets, using well-established interpretability techniques such as SHAP and pruned decision trees, demonstrated strong convergence between our results and those produced by standard baselines. These findings highlight the promise of causal-argumentative explanations as a complementary perspective within the broader landscape of interpretable machine learning. Future work will expand this approach along several dimensions. First, we will explore additional argumentation semantics and acceptance criteria to assess their impact on stability and expressiveness. Second, we intend to evaluate the method on more complex datasets and higher-dimensional feature spaces. Third, we plan to introduce quantitative evaluation metrics, including fidelity measures that assess how faithfully the explanations reflect model behavior, counterfactual consistency tests that verify whether explanations change appropriately under feature interventions, and robustness evaluations under distributional shift. Finally, we will investigate strategies to reduce computational overhead, enabling broader scalability and practical deployment in real-world systems.

\section{Use of Generative AI}

Generative AI tools were used to assist in LaTeX formatting, coding error analysis, and writing text improvement. All AI-generated outputs were reviewed, verified, and edited by the authors to ensure accuracy and integrity.

\bibliographystyle{splncs04}
\bibliography{references}

\end{document}